\documentclass{article}

\PassOptionsToPackage{numbers, compress}{natbib}

 \usepackage[preprint, final]{neurips_2026}

\usepackage[utf8]{inputenc} 
\usepackage[T1]{fontenc}    
\usepackage{hyperref}       
\usepackage{url}            
\usepackage{booktabs} 
\usepackage{pifont}
\usepackage{multirow}       
\usepackage{tabularx}       
\usepackage{amsmath}        
\usepackage{amsfonts}       
\usepackage{amssymb}        
\usepackage{graphicx}
\usepackage{nicefrac}       
\usepackage{microtype}      
\usepackage{xcolor}         

\newcommand{\sys}{MoE-XBench}

\title{Benchmarking Composable Compression Techniques in Mixture-of-Experts LLMs
}

\author{%
Afsara Benazir\thanks{Work conducted during an internship at Sony AI} \\
  University of Virginia\\
  \texttt{hys4qm@virginia.edu}
\And
  Chen Chen\\
  Sony AI \\
  \texttt{chena.chen@sony.com} 
\And
  Rongxiao Qu\\
  Sony AI \\
  \texttt{Rongxiao.Qu@sony.com} 
\And
  Jiabo Huang\\
  Sony AI \\
  \texttt{raymond.huang@sony.com} 
\And
  Jingtao Li\\
  Sony AI \\
  \texttt{jingtao.li@sony.com} 
\And
  Lingjuan Lyu \\
  Sony AI \\
  \texttt{lingjuan.lv@sony.com} 
}

\begin{document}

\maketitle
\begin{abstract}
Mixture-of-Experts (MoE) LLMs scale model capacity efficiently through sparse activation, but their large expert parameter footprint, routing imbalance, and long-context KV-cache growth make deployment difficult on commodity hardware. Practical deployment often requires stacking multiple compression techniques: expert pruning removes redundant experts, weight quantization lowers model memory footprint, and KV-cache compression reduces long-context memory pressure. However, these techniques are typically evaluated in isolation, leaving open how they interact when applied together in realistic deployment pipelines.
In this work, we present \sys{}, a systematic benchmark for evaluating composable MoE compression as an end-to-end deployment workflow. \sys{} studies 10 MoE models ranging from 30B to 235B parameters across standard-attention, hybrid linear-attention, and sliding-window attention architectures. Across seven workloads, it evaluates 20\%-50\% expert pruning rates, 1 to 16 bit weight-quantization schemes, and multiple KV-cache precision settings, applied both individually and in combination.
\sys{} introduces an eight-module evaluation suite that jointly measures composable compression quality, workload and architecture robustness, pruning/quantization/KV cache sensitivity, and deployment efficiency on commodity hardware. Our results reveal non-trivial interactions among compression methods: composable compression cannot be predicted from standalone techniques, compression rate alone does not reliably predict quality loss or runtime gain, expert pruning is the dominant degradation source, and average quality can hide workload and architecture-specific failures. By releasing normalized module scores, compressed artifacts, and reproducible scripts, \sys{} enables practical accuracy-memory-latency comparison across MoE model families and hardware backends.
\end{abstract}

\section{Introduction}
Large language models (LLMs) achieve strong performance across reasoning, coding, instruction-following, and long-context tasks, but their large parameter counts make deployment expensive in memory, latency, and energy. Mixture-of-Experts (MoE) LLMs improve this tradeoff by routing each token to only a small subset of experts, thereby scaling total model capacity while keeping active computation relatively small. However, MoE models remain difficult to deploy on commodity hardware because they require storing many experts, managing expert-prefetching bottlenecks, handling expert imbalance, and maintaining a large KV cache during long-context inference that can introduce non-trivial accuracy-efficiency tradeoffs.
Existing LLM compression methods remain far from practical deployment because of these key challenges:

\begin{figure*}
    \centering
    \includegraphics[width=0.75\linewidth]{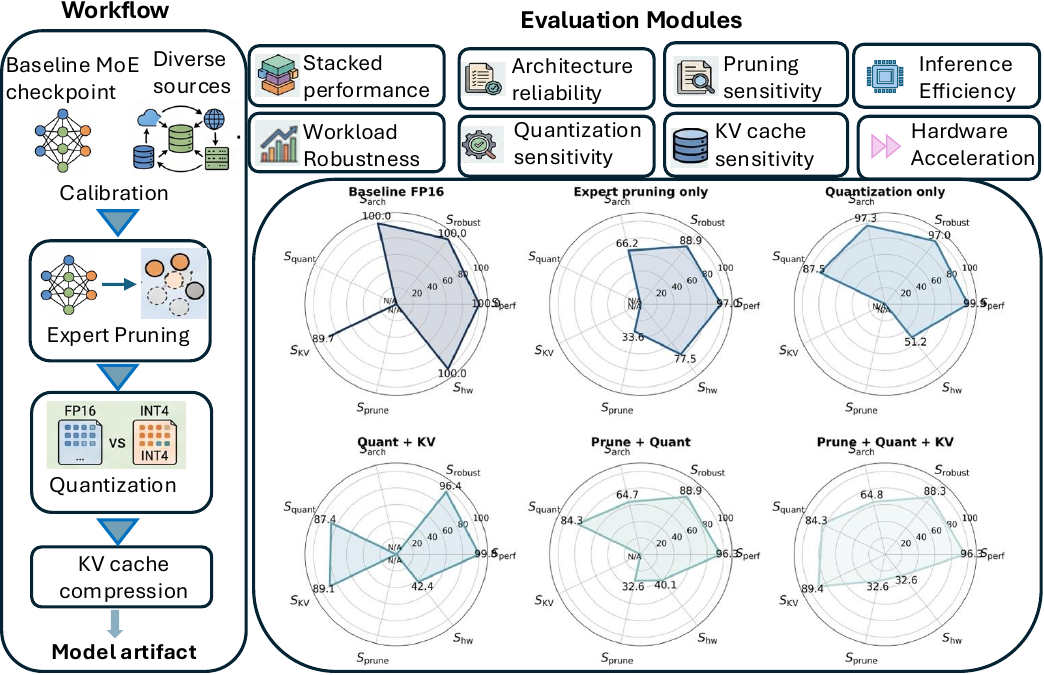}
    \caption{Overview of \sys{}. Each radar reports normalized module scores, where higher is better. Performance, robustness, and reliability measure fixed-configuration quality, while pruning, quantization, and KV tolerance summarize controlled sensitivity sweeps around that configuration.
    }
    \label{fig:o}
    \vspace{-2mm}
\end{figure*}
\textbf{(1) Lack of systematic evaluation of MoE compression schemes.} Prior work has extensively studied compression for dense LLMs~\cite{zhu2024survey,  yang2024llmcbench, frantar2022gptq, liu2024kivi}, but the same has not been done for MoE LLMs, where sparse routing, expert imbalance, and expert redundancy create different accuracy-efficiency tradeoffs~\cite{hu2025moequant}. These effects further vary across MoE architectures - standard-attention~\cite{jiang2024mixtral}, hybrid linear-attention~\cite{lieber2024jamba, kimi2025linear}, and sliding-window attention~\cite{step} MoEs.

\textbf{(2) Lack of prior work on composable compression in MoE LLMs.} Existing compression studies largely evaluate pruning, quantization, and KV-cache compression as isolated techniques~\cite{lasby2025reap, frantar2023qmoe, hu2025moequant, liu2024kivi, dehghanighobadi2026depthkv}. Yet stacking them introduces non-obvious interactions: pruning changes expert utilization, quantization changes numerical error, and KV-cache compression changes attention behavior. It remains unclear whether composable effects are additive, redundant, or harmful, leaving practitioners to rely on heuristic choices of pruning ratio, quantization format and KV-cache precision.

\textbf{(3) Existing efficiency metrics do not reflect real deployment cost.} A major challenge in evaluating LLM compression is that efficiency is often measured using proxy metrics such as parameter count, FLOPs etc. which do not fully capture runtime behavior. A smaller MoE model may not yield proportional speedups: inference also depends on context length, batch size, backend kernels, memory layout, dequantization overhead, and hardware scheduling. Evaluation should therefore verify whether compression improves runtime metrics such as peak memory, prefill/decode throughput, and hardware acceleration on target device.

To address aforementioned challenges, we present 
\sys{}, 
a comprehensive benchmark for systematic evaluation of compression methods in Mixture-of-Experts (MoE) LLMs. We design an end-to-end workflow that converts the original MoE checkpoints into deployable artifacts through calibration, expert pruning, weight quantization, and KV-cache compression (ref. \autoref{fig:o}).
\sys{} analyzes MoE deployment tradeoffs across architectures and commodity hardware, evaluating 10 MoE LLMs on seven workloads under varying pruning rate, quantization, and KV-cache precision. Our main contributions are as follows:
\vspace{-2ex}

\begin{itemize}  
\item We build \sys{}, an end-to-end benchmarking framework for MoE compression that frames expert pruning, weight quantization, and KV-cache compression as components of a composable deployment pipeline, establishing the first systematic study of composable compression in MoE models.

\item We design an evaluation suite consisting of eight modules to address the research questions in \autoref{tab:rq} 
separating model quality from deployment efficiency while characterizing nontrivial interactions among individual and composable compression techniques across model accuracy, workload type, architecture robustness, efficiency, and hardware backend. 

\item We release a reproducible benchmark implementation and empirical study across MoE families 
on commodity hardware, identifying practical deployment tradeoffs that show when compression preserves quality, where losses compound, and whether model-size reductions translate into real gains in memory, throughput, latency, and hardware efficiency.


\end{itemize}

\begin{table}
    \centering
    \includegraphics[width=\linewidth]{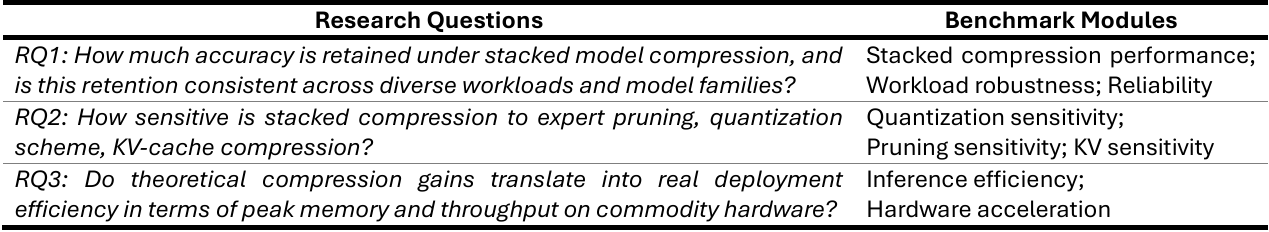}
    \caption{Research questions and their corresponding \sys{} evaluation modules.
    }
    \label{tab:rq}
\end{table}

\textit{\textbf{Our takeaway.}
By extensively benchmarking MoE models across diverse architectures, compression rates, and hardware backends, we find that MoE compression must be evaluated as a deployment pipeline rather than as isolated accuracy-preserving steps. We observe that model quality is dependent on compression axis, architecture, and is non-additive, with expert pruning dominating degradation. Hardware efficiency is often decoupled from memory reduction due to KV-cache and low-bit dequantization overheads. These observations underscore the practical value of \sys{} in identifying the compression point that preserves quality while delivering real throughput and memory gains.}




\section{Background}
\label{sec:b}
\subsection{Mixture-of-Experts LLMs}
Mixture-of-Experts (MoE) LLMs~\citep{shazeer2017moe} route each token to a small subset of experts, increasing total parameter capacity without proportionally increasing active computation~\citep{fedus2022switch}. This sparsity improves capacity-efficiency tradeoffs but introduces deployment challenges such as routing overhead, expert imbalance, and large memory footprints. Modern MoE LLMs vary in routing design: Qwen3-MoE~\citep{qwen3moe} and GLM-4.5~\citep{glm45} use standard sparse top-$k$ routing, while DeepSeekMoE~\citep{dai2024deepseekmoe} adds shared experts to reduce redundancy. Because these choices change routing patterns, active parameters, and memory/throughput behavior, compression results should be validated across families rather than assumed to transfer directly.

\subsection{Model Compression}
Compression for MoE LLM deployment spans a wide design space; we study three post training compression axes that act directly on the deployed model and its inference-time state: offline expert pruning and weight quantization, and runtime KV-cache compression. Other techniques are complementary but either require retraining (distillation~\citep{kim2025every}, merging~\citep{jha2026ream}) or live at the serving layer (offloading~\citep{eliseev2023moeoffload}, speculative decoding~\citep{leviathan2023speculative}); we treat them as orthogonal.


\textbf{Expert pruning} removes low-utility experts while preserving sparse routing, with REAP~\citep{lasby2025reap} showing its effectiveness as one-shot MoE compression. 
\textbf{Quantization} lowers weight precision to reduce memory footprint; QMoE demonstrates sub-1-bit effective MoE storage~\citep{frantar2023qmoe}, while MoEQuant uses expert-aware calibration for sparse-routing imbalance~\citep{hu2025moequant}. 
\textbf{KV-cache compression} reduces long-context memory growth, complementing weight compression through methods such as KIVI’s asymmetric 2-bit quantization~\citep{liu2024kivi}, layer-dependent pruning~\citep{dehghanighobadi2026depthkv}, and TurboQuant~\citep{zandieh2025turboquant}.

\subsection{Gap in Prior Work}
Dense LLM compression has been studied extensively, with a broad literature on pruning, low-bit quantization, and inference-time memory reduction~\citep{zhu2024survey}. MoE compression is much less systematically characterized: existing studies typically focus on a limited number of models, one compression axis at a time, and narrow evaluation settings, making it difficult to separate architecture-specific observations from general deployment trends~\citep{lasby2025reap, frantar2023qmoe, hu2025moequant, liu2024kivi, dehghanighobadi2026depthkv}. More importantly, no prior work treats composable compression as a first-class evaluation setting for MoEs. In practice, deployable models combine expert pruning, weight quantization, and KV-cache compression in a pipeline, yet the interactions among these stages remain poorly understood: because each acts on a different part of the inference stack, it is unclear whether composable compression is complementary, redundant, or harmful for MoE models, and whether theoretical savings translate into real gains in memory and throughput on target hardware.


LLMCBench~\cite{yang2024llmcbench} evaluates weight pruning and quantization
schemes separately for dense model architectures, 
but does not consider composable compression pipelines, KV-cache compression, or MoE architectures.
MoE-CAP~\cite{jiang2024moecap}, the closest MoE benchmark, targets serving utilization rather than model compression and does not study these techniques as composable. Method-specific studies - REAP~\cite{lasby2025reap}, QMoE~\cite{frantar2023qmoe}, KIVI~\cite{liu2024kivi} etc. evaluate compression along a single axis with one or two metric on one architecture family. None jointly evaluate all three MoE-relevant compression axes, study them as a composable pipeline, or span multiple MoE architecture families. \sys{} fills this gap by evaluating all combinations of pruning, quantization, and KV-cache compression across three MoE architecture families (standard, hybrid linear-attention, sliding-window-attention) through an eight-module suite that covers quality retention, robustness, sensitivity, and runtime efficiency.

\section{\sys{}}
\subsection{Overview}
\sys{} is an evaluation suite for benchmarking compression choices in MoE LLMs. It supports both standalone techniques and combined compression pipelines, including expert pruning, weight quantization, and KV-cache compression. \sys{} is necessary to evaluate MoE-specific compression axes and their interactions under a unified methodology and is distinct from other compression benchmarks such as \cite{yang2024llmcbench, jiang2024moecap}.

\noindent\textbf{Workflow.}
As shown in \autoref{fig:o}, \sys{} uses a one-shot compression pipeline to transform baseline MoE checkpoints into deployable compressed artifacts.
Each resulting artifact is evaluated using a unified harness organized around
these modules (\S\ref{subsec:benchmark_modules}) that report both accuracy and efficiency, enabling controlled comparison between individual compression stages and combined compression pipelines. 
We evaluate the compression configurations in \autoref{tab:compression_configs}.

\subsection{Design Goals}
\label{subsec:design_goals}

\sys{} is built around three design goals.

\textbf{G1: Compression methods as composable building blocks.} Expert pruning, weight quantization, and KV-cache compression are treated as deployment primitives that can be applied individually or combined. This design reflects how large MoE models are prepared for practical deployment under memory and latency constraints, while enabling controlled comparison between standalone and combined compression settings.

\textbf{G2: Separation of model quality from deployment efficiency.} A compressed MoE model is useful only if it preserves task quality while reducing practical deployment cost. \sys{} reports both model accuracy and efficiency metrics, so that compression gains in one dimension are not obscured by losses in another.

\textbf{G3: Expose interactions between compression methods.}
Composable compression may not behave additively: expert pruning can change quantization tolerance, quantization can interact with calibration quality, and KV-cache compression can affect pruned and dense models differently. \sys{} includes dedicated modules to quantify individual and combined effects, isolating each compression choice while exposing how composable pipelines affect evaluation metrics.


\begin{table}[t]
\centering
\footnotesize
\caption{Compression configurations in \sys{}; the last three are composable.}
\label{tab:compression_configs}
\setlength{\tabcolsep}{4pt}
\renewcommand{\arraystretch}{1.1}
\resizebox{\textwidth}{!}{%
\begin{tabular}{lcccccc}
\toprule
\textbf{Configuration} & \textbf{Baseline FP16} & \textbf{Prune only} & \textbf{Quant only} & \textbf{Quant + KV} & \textbf{Prune + Quant} & \textbf{Prune + Quant + KV} \\
\midrule
\textbf{Expert Pruning}       & {\color{red}\ding{55}} & {\color{green!60!black}\ding{51}} & {\color{red}\ding{55}} & {\color{red}\ding{55}} & {\color{green!60!black}\ding{51}} & {\color{green!60!black}\ding{51}} \\
\textbf{Weight Quantization}  & {\color{red}\ding{55}} & {\color{red}\ding{55}} & {\color{green!60!black}\ding{51}} & {\color{green!60!black}\ding{51}} & {\color{green!60!black}\ding{51}} & {\color{green!60!black}\ding{51}} \\
\textbf{KV Cache Compression} & {\color{red}\ding{55}} & {\color{red}\ding{55}} & {\color{red}\ding{55}} & {\color{green!60!black}\ding{51}} & {\color{red}\ding{55}} & {\color{green!60!black}\ding{51}} \\
\bottomrule
\end{tabular}%
}
\end{table}
\subsection{Benchmark Modules}
\label{subsec:benchmark_modules}

\sys{} contains eight evaluation modules. Each reports raw measurements and a normalized score $\mathrm{Score}_x$, where 100 denotes parity with a reference setting and higher is better. Accuracy scores average over models and datasets; efficiency scores average over models, hardware platforms, and runtime settings. We use raw measurements for detailed analysis and normalized scores for compact comparison and Pareto analysis.
For retention-stability scores, we use
\[
\Gamma(\{x_i\})=\operatorname{GM}_i(x_i)\cdot \frac{\min_i x_i}{\max_i x_i},
\]
which rewards high average retention and penalizes uneven degradation.
For perplexity-based modules (Modules 3-6 on WikiText), we define $\widetilde{S}=1/\mathrm{PPL}$ 
so that higher $\widetilde{S}$ corresponds to better quality, and retention ratios $R$ remain in $(0,1]$ when compression hurts. In all sensitivity modules, the reference-format set is excluded from the sweep set: $\mathcal{B}^{-}=\mathcal{B}\setminus\{b_0\}$, $\mathcal{K}^{-}=\mathcal{K}\setminus\{K_0\}$, and $\mathcal{P}^{-}=\mathcal{P}\setminus\{p_0\}$.

\textbf{Module 1: Stacked compression performance.}
This module measures task quality retained after composable compression. Let $B$ be the full-precision base model and $c$ a compressed configuration. If $S_c$ and $S_B$ are mean task scores, then
\[
\mathrm{Score}_{\mathrm{perf}}(c)=100\cdot \frac{S_c}{S_B}.
\]

\textbf{Module 2: Workload robustness.}
This module measures whether quality is retained consistently across workloads. For workload $w\in\mathcal{W}$, let $R_{c,w}=S_{c,w}/S_{B,w}$. We define
\[
\mathrm{Score}_{\mathrm{robust}}(c)=100\cdot \Gamma(\{R_{c,w}:w\in\mathcal{W}\}).
\]

\textbf{Module 3: Architecture reliability.}
This module measures whether a compression setup remains reliable across MoE architecture families. For model $m$, let $R_{c,m}=S_{c,m}/S_{B,m}$, and for family $a$, let $A_{c,a}=\operatorname{GM}_{m\in a}(R_{c,m})$. We define
\[
\mathrm{Score}_{\mathrm{arch}}(c)=100\cdot \Gamma(\{A_{c,a}\}_a).
\]

\textbf{Module 4: Quantization sensitivity.}
This module measures quality retained as weight precision is reduced. Let $b_0$ denote the
16-bit reference and $\mathcal{B}^{-}$ the evaluated low-bit formats. For configuration $c$,
\[
R_{c,b}^{\mathrm{quant}}
=
\frac{\widetilde{S}_{c,b}}{\widetilde{S}_{c,b_0}},
\qquad
\mathrm{Score}_{\mathrm{quant}}(c)
=
100 \times
\mathrm{GM}_{b \in \mathcal{B}^{-}}\!\left(R_{c,b}^{\mathrm{quant}}\right)
\cdot
\frac{\min_{b \in \mathcal{B}^{-}} R_{c,b}^{\mathrm{quant}}}
{\max_{b \in \mathcal{B}^{-}} R_{c,b}^{\mathrm{quant}}}.
\]

\textbf{Module 5: KV-cache sensitivity.}
This module measures quality retained as KV-cache precision is reduced. Let $K_0$ denote the
fp16/fp16 KV-cache reference and $\mathcal{K}^{-}$ the compressed KV formats. For configuration $c$,
\[
R_{c,k}^{\mathrm{KV}}
=
\frac{\widetilde{S}_{c,k}}{\widetilde{S}_{c,K_0}},
\qquad
\mathrm{Score}_{\mathrm{KV}}(c)
=
100 \times
\mathrm{GM}_{k \in \mathcal{K}^{-}}\!\left(R_{c,k}^{\mathrm{KV}}\right)
\cdot
\frac{\min_{k \in \mathcal{K}^{-}} R_{c,k}^{\mathrm{KV}}}
{\max_{k \in \mathcal{K}^{-}} R_{c,k}^{\mathrm{KV}}}.
\]

\textbf{Module 6: Expert pruning sensitivity.}
This module measures quality retained as experts are removed. Let $p_0$ denote the unpruned
reference and $\mathcal{P}^{-}$ the evaluated pruning settings. For configuration $c$,
\[
R_{c,p}^{\mathrm{prune}}
=
\frac{\widetilde{S}_{c,p}}{\widetilde{S}_{c,p_0}},
\qquad
\mathrm{Score}_{\mathrm{prune}}(c)
=
100 \times
\mathrm{GM}_{p \in \mathcal{P}^{-}}\!\left(R_{c,p}^{\mathrm{prune}}\right)
\cdot
\frac{\min_{p \in \mathcal{P}^{-}} R_{c,p}^{\mathrm{prune}}}
{\max_{p \in \mathcal{P}^{-}} R_{c,p}^{\mathrm{prune}}}.
\]
\textbf{Module 7: Inference efficiency.}
This module measures deployment-cost reduction. Let $M_c$, $T_{c,\mathrm{prefill}}$, and $T_{c,\mathrm{decode}}$ be peak memory, prefill latency, and decode latency for configuration $c$, with base-model values subscripted by $B$. Raw values are reported separately. We define
\[
\mathrm{Score}_{\mathrm{cost}}(c)=100\cdot \operatorname{GM}\left(\frac{M_B}{M_c},\frac{T_{B,\mathrm{prefill}}}{T_{c,\mathrm{prefill}}},\frac{T_{B,\mathrm{decode}}}{T_{c,\mathrm{decode}}}\right).
\]


\textbf{Module 8: Hardware acceleration.}
This module measures effective runtime speedup per unit effective memory compression. Let
$S_c=\operatorname{GM}\left(\frac{T_{B,\mathrm{prefill}}}{T_{c,\mathrm{prefill}}},\frac{T_{B,\mathrm{decode}}}{T_{c,\mathrm{decode}}}\right)$
and
$CR^{\mathrm{eff}}_c=\frac{W_B+KV_B}{W_c+KV_c}$.
Here, $B$ is the baseline, $c$ is the compressed configuration, $T$ is phase latency, $W$ is weight memory, and $KV$ is KV-cache memory at the same context length and batch size. We define
\[
\mathrm{Score}_{\mathrm{hw}}(c)=100\cdot \frac{S_c}{CR^{\mathrm{eff}}_c}.
\]

A score below 100 means compression does not translate into proportional speedup. Note that this score measures speedup relative to compression rate, not absolute speedup.

\vspace{-1ex}
\section{Experimental Setup}
\vspace{-1ex}
\label{sec:experimental_setup}

\textbf{Models.}
We evaluate 10 MoE LLMs across standard-attention, hybrid linear-attention, and sliding-window-attention families 
: Qwen3-30B, Qwen3-Coder-30B, Qwen3-235B, GLM-4.7-Flash, MiniMax-M2, Kimi-Linear-48B, Qwen3-Next-80B, Qwen3-Coder-Next, Qwen3.6-35B, and Step-3.5-Flash. These models span diverse scales, attention architecture, and deployment use cases. Modules 1-2 and 4-6 use the primary Qwen3 models for controlled compression sweeps, while Module 3 uses the full set for architecture reliability.


\textbf{Compression Settings.}
We evaluate the six configurations in \autoref{tab:compression_configs}, covering empty, single-axis, and composable subsets of expert pruning, weight quantization, and KV-cache compression. We use representative methods for each orthogonal axis: REAP~\citep{lasby2025reap} for expert pruning, GGUF quantization \cite{llama} for weight and uniform/turboquant KV-cache compression. evaluation with alternate expert compression method (REAM \cite{jha2026ream}) in \autoref{subsec:appendix-ream}.
Unless otherwise stated, 
the default setting is 20\% pruning, Q4\_K\_M model precision, and Q8 KV cache. All reported latency/throughput measurements use BS=1, a deliberate choice: \sys{} targets commodity-hardware, local deployment, where single-request serving dominates. Sensitivity studies sweep pruning ratios from 25\%-50\%, model weight precision from 1 to 16 bits, and KV formats including FP16, Q8, Q4, and TurboQuant. We use the same four calibration datasets as REAP~\citep{lasby2025reap}.


\textbf{Tasks and Datasets.}
We evaluate seven workload categories that are central to real-world use cases: knowledge (MMLU~\citep{hendrycks2021mmlu}), math reasoning (GSM8K~\citep{cobbe2021gsm8k}), generative reasoning (MuSR~\citep{sprague2024musr}), instruction following (IFEval~\citep{zhou2023ifeval}), code generation (HumanEval~\citep{chen2021humaneval}), tool use (BFCLv3~\citep{bfclv3}), and long-context understanding (RULER~\citep{hsieh2024ruler}).
Modules 3-6 report perplexity on Wikitext-2~\citep{merity2017wikitext}.


\textbf{Hardware Setup.}
We evaluate efficiency on an Apple M1 Max with 64 GB unified memory, NVIDIA H100 with 80 GB RAM 
covering unified and non-unified memory architectures. We use llama.cpp \cite{llama} \texttt{[build b9050]} for inference, while keeping the evaluation framework-agnostic as GGUF support expands across backends such as vLLM~\citep{kwon2023vllm} and SGLang~\citep{zheng2024sglang}. We use \texttt{llama-cpp-turboquant} as reference implementation for TurboQuant.




\textbf{Measurement Metrics.}
We report raw measurements and the normalized module scores defined in \S\ref{subsec:benchmark_modules}. Accuracy metrics include task accuracy and perplexity. Efficiency metrics include compression rate, peak memory, prefill and decode throughput.
Raw results are reported per model, dataset, hardware, context length, batch size, and KV-cache setting; normalized scores summarize cross-configuration tradeoffs across compression configurations.


\vspace{-1ex}
\section{Accuracy Evaluation}
\vspace{-1ex}

\begin{table}[t]
    \centering
    \caption{Performance comparison across different compression configurations for Qwen3-30B-A3B-Instruct-2507 (Q-30B-A3B) and Qwen3.6-35B-A3B (Q-35B-A3B) on diverse workloads.}
    \label{tab:ae}
    \setlength{\tabcolsep}{3pt}
    \renewcommand{\arraystretch}{1.1}
    \resizebox{\linewidth}{!}{%
    \begin{tabular}{llc cccccccc cc}
        \toprule
        \multirow{2}{*}{\textbf{Method}} & \multirow{2}{*}{\textbf{Model}} & \multirow{2}{*}{\shortstack{\textbf{Model Size}\\ \textbf{(Compression Rate)}}} & \multicolumn{8}{c}{\textbf{Workload}} & \multirow{2}{*}{\textbf{Score}$_{\text{perf}}$} & \multirow{2}{*}{\textbf{Score}$_{\text{robust}}$} \\
        \cmidrule(lr){4-11}
        & & & GSM8K & MMLU & MuSR & IFEval & HumanEval & BFCLv3 & RULER & Avg. & & \\
        \midrule
        \multirow{2}{*}{Baseline (BF16)}
            & Q-30B-A3B & 61.10 GB (0.0\%)  & 0.8360 & 0.8470 & 0.4290 & 0.8170 & 0.9390 & 0.8880 & 0.9441 & 0.8143 & \multirow{2}{*}{100.00} & \multirow{2}{*}{100.00} \\
            & Q-35B-A3B & 69.38 GB (0.0\%)  & 0.8741 & 0.8657 & 0.4405 & 0.8207 & 0.9695 & 0.8662 & 0.9470 & 0.8262 & & \\
        \midrule
        \multirow{2}{*}{Prune only}
            & Q-30B-A3B & 44.00 GB (28.0\%) & 0.8400 & 0.7420 & 0.4380 & 0.7650 & 0.8110 & 0.8560 & 0.9452 & 0.7710 & \multirow{2}{*}{96.96}  & \multirow{2}{*}{88.87}  \\
            & Q-35B-A3B & 52.04 GB (25.0\%) & 0.8779 & 0.8467 & 0.4444 & 0.8022 & 0.9634 & 0.8554 & 0.9473 & 0.8196 & & \\
        \midrule
        \multirow{2}{*}{Quant. only}
            & Q-30B-A3B & 18.63 GB (69.5\%) & 0.8330 & 0.8450 & 0.4250 & 0.8260 & 0.9270 & 0.8910 & 0.9422 & 0.8127 & \multirow{2}{*}{99.91}  & \multirow{2}{*}{97.00}  \\
            & Q-35B-A3B & 21.39 GB (69.2\%) & 0.8711 & 0.8655 & 0.4392 & 0.8392 & 0.9573 & 0.8644 & 0.9477 & 0.8263 & & \\
        \midrule
        \multirow{2}{*}{Quant. + KV}
            & Q-30B-A3B & 18.63 GB (69.5\%) & 0.8378 & 0.8422 & 0.4246 & 0.8336 & 0.9268 & 0.8694 & 0.9435 & 0.8111 & \multirow{2}{*}{99.77}  & \multirow{2}{*}{96.42}  \\
            & Q-35B-A3B & 21.39 GB (69.2\%) & 0.8681 & 0.8655 & 0.4365 & 0.8336 & 0.9512 & 0.8775 & 0.9473 & 0.8257 & & \\
        \midrule
        \multirow{2}{*}{Prune + Quant.}
            & Q-30B-A3B & 13.42 GB (78.0\%) & 0.8230 & 0.7330 & 0.4310 & 0.7430 & 0.8170 & 0.8410 & 0.9424 & 0.7615 & \multirow{2}{*}{96.35}  & \multirow{2}{*}{88.91}  \\
            & Q-35B-A3B & 16.04 GB (76.9\%) & 0.8779 & 0.8443 & 0.4378 & 0.8133 & 0.9390 & 0.8769 & 0.9447 & 0.8191 & & \\
        \midrule
        \multirow{2}{*}{Prune + Quant. + KV}
            & Q-30B-A3B & 13.42 GB (78.0\%) & 0.8180 & 0.7320 & 0.4270 & 0.7600 & 0.8050 & 0.8420 & 0.9435 & 0.7611 & \multirow{2}{*}{96.31}  & \multirow{2}{*}{88.28}  \\
            & Q-35B-A3B & 16.04 GB (76.9\%) & 0.8772 & 0.8447 & 0.4418 & 0.8115 & 0.9390 & 0.8729 & 0.9453 & 0.8189 & & \\
        \bottomrule
    \end{tabular}%
    }
\end{table}


\begin{table}[t]
    \centering
    \caption{Cross-architecture reliability across compression config. PPL, $\downarrow$ is better; score, $\uparrow$ is better.
    }
    \label{tab:r}
    \resizebox{\linewidth}{!}{%
    \begin{tabular}{lcccccc}
        \toprule
        \textbf{Model} & \textbf{Baseline} & \textbf{Pruning} & \textbf{Quant only} & \textbf{Quant+KV} & \textbf{Prune+Quant} & \textbf{Prune+Quant+KV} \\
        \midrule
        \multicolumn{7}{c}{\textit{Standard MoE}} \\
        \midrule
        Qwen3-30B-A3B-Instruct          & 7.36   & 9.46   & 7.41   & 7.49   & 9.55   & 9.54   \\
        Qwen3-Coder-30B-A3B-Instruct    & 9.54   & 12.88  & 9.74   & 9.71   & 13.11  & 13.11  \\
        Qwen3-235B-A22B-Instruct-2507   & 4.31   & 6.50   & 4.41   & 4.41   & 6.62   & 6.62   \\
        GLM-4.7-Flash                   & 9.86   & 17.64  & 10.27  & 9.85   & 18.25  & 18.25  \\
        MiniMax-M2                      & 6.79   & 13.16  & 7.09   & 7.13   & 10.41  & 10.41  \\
        \midrule
        \textit{Aggregate (Score$_{\text{arch}}$)} & 100.00  & 65.13   & 97.35   & 97.90   & 66.85   & 66.86   \\
        \midrule
        \multicolumn{7}{c}{\textit{Hybrid Linear-Attn MoE}} \\
        \midrule
        Kimi-Linear-48B-A3B-Instruct    & 7.08   & 9.63   & 7.17   & 7.17   & 9.79   & 9.79   \\
        Qwen3-Next-80B-A3B-Instruct     & 5.72   & 7.37   & 5.82   & 5.82   & 7.47   & 7.48   \\
        Qwen3-Coder-Next                & 8.23   & 13.74  & 8.35   & 8.35   & 13.94  & 13.95  \\
        Qwen3.6-35B-A3B           & 6.72   & 8.67   & 6.82   & 6.82   & 8.80   & 8.80   \\
        \midrule
        \textit{Aggregate (Score$_{\text{arch}}$)} & 100.00  & 72.15   & 98.53   & 98.55   & 71.09   & 71.06   \\
        \midrule
        \multicolumn{7}{c}{\textit{Sliding-window Attention MoE}} \\
        \midrule
        Step-3.5-Flash                  & 2.27   & 5.40   & 2.57   & 2.57   & 6.27   & 6.26   \\
        \midrule
        \textit{Overall Score$_{\text{arch}}$}     & 100.00  & 65.63   & 96.93   & 97.21   & 65.48   & 65.48   \\
        \bottomrule
    \end{tabular}%
    }
\end{table}

\begin{figure}[t]
    \centering

    \begin{minipage}[t]{0.48\linewidth}
        \centering
        \includegraphics[width=\linewidth]{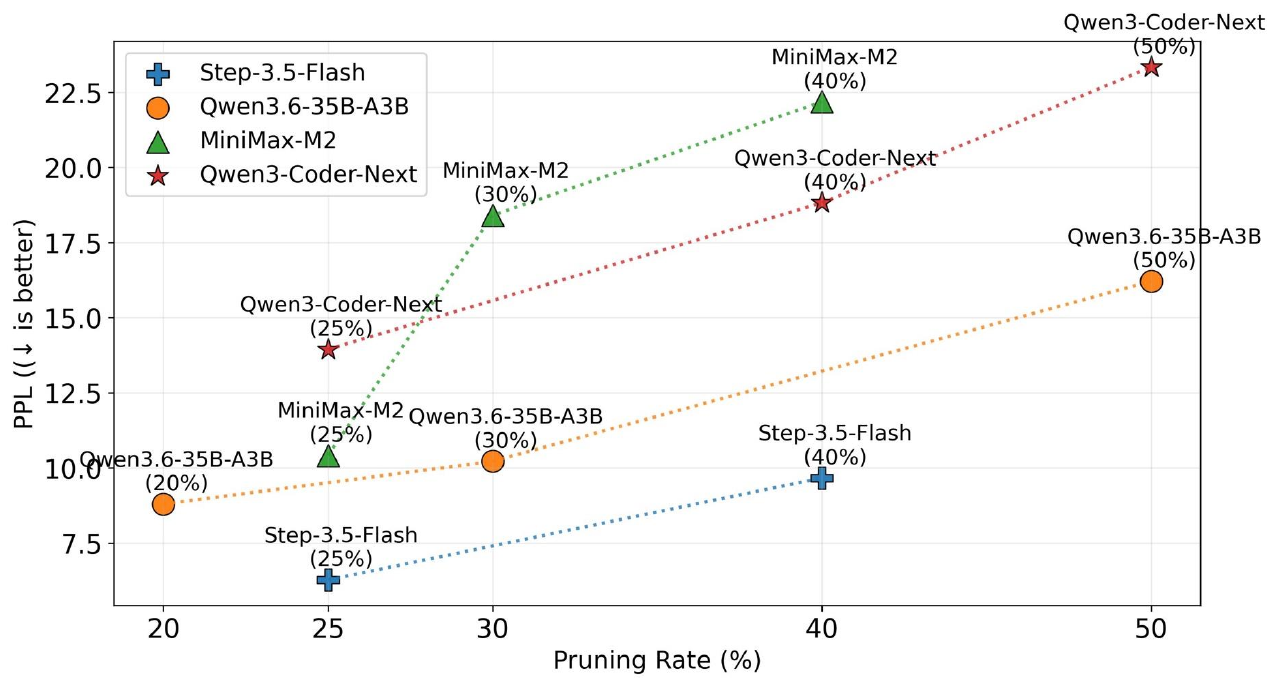}
        \caption{Model performance at different pruning rates validate finding 2.
        }
        \label{fig:ps}
    \end{minipage}
    \hfill
    \begin{minipage}[t]{0.48\linewidth}
        \centering
        \includegraphics[width=\linewidth]{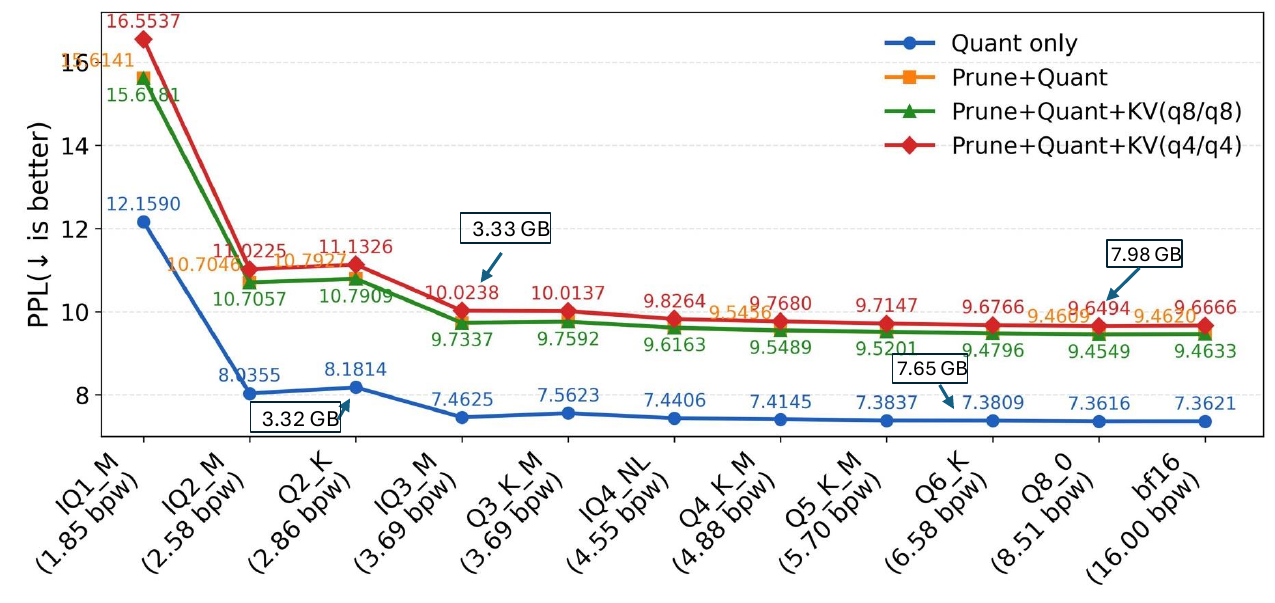}
        \caption{Quantization sensitivity for Qwen3-30B-A3B across bit-widths.}
        \label{fig:qs}
    \end{minipage}

\end{figure}
\begin{table}[t]
    \centering
    \caption{KV-cache sensitivity.
Higher $\mathrm{score}_{kv}$ indicates
greater tolerance to KV-cache compression.}
    \label{tab:kvcs}
    \setlength{\tabcolsep}{4pt}
    \renewcommand{\arraystretch}{1.15}
    \resizebox{0.9\linewidth}{!}{%
    \begin{tabular}{lccccccc}
        \toprule
        \textbf{Configuration} & \textbf{Fp16/fp16} & \textbf{Q8/Q8} & \textbf{Q4/Q4} & \textbf{Q8/Turbo4} & \textbf{Q8/Turbo3} & \textbf{Turbo4/Turbo3} & \textbf{score$_{\mathrm{kv}}$} \\
        \midrule
        Baseline FP16        & 6.4935 & 6.4949 & 6.6130 & 6.5127 & 6.5311 & 6.7846 & \textbf{94.38} \\
        Quant + KV           & 6.5509 & 6.5492 & 6.6741 & 6.5614 & 6.5795 & 6.8459 & \textbf{94.37} \\
        Prune + KV           & 8.1727 & 8.1730 & 8.3441 & 8.2070 & 8.2417 & 8.6205 & \textbf{93.18} \\
        Prune + Quant + KV   & 8.2479 & 8.2469 & 8.4371 & 8.2856 & 8.3184 & 8.7095 & \textbf{93.00} \\
        \bottomrule
    \end{tabular}%
    }
\end{table}

We attempt to answer RQ1 and RQ2 through this empirical study and primarily look into the model quality retention, robustness and interaction (composable effect) scores (Module 1-6). 

\textbf{Finding 1: Compression ratio is not predictive of quality loss.}
We challenge the common assumption that quality degradation scales with the fraction of parameters or bits removed. On Qwen3-30B-A3B, 19.35\% expert pruning increases PPL by 28.5\%, while 69.5\% Q4\_K\_M weight quantization increases PPL by only 0.7\% (\autoref{fig:qs}). 
The same asymmetry appears in accuracy: 
prune-only obtains $\mathrm{Score}_{\mathrm{perf}}=96.96$ and $\mathrm{Score}_{\mathrm{robust}}=88.87$, while Prune+Quant remains close at 96.35/88.91 despite substantially higher compression; adding KV-Q8 further changes the scores only marginally to 96.31/88.28 (\autoref{tab:ae}). The PPL gap between Prune+Quant and Prune+Quant+KV is negligible across models (\autoref{tab:r}). 
Thus, the compression axis matters more than the nominal compression rate: a smaller pruning step ($\sim$25\%) hurts more than aggressive 4-bit quantization.

\textbf{Finding 2: MoE compression is architecture-dominated, not scale-dominated.}
Larger expert pools do not necessarily imply greater pruning tolerance. Qwen3-235B-A22B suffers a 54\% PPL increase under Prune+Quant, while the 6.71x smaller Qwen3.6-35B-A3B increases by only 31\% (\autoref{tab:r}). Across pruning sweeps, MiniMax-M2 remains highly sensitive despite its size, whereas Step-3.5-Flash and Qwen3.6-35B-A3B degrade more gradually (\autoref{fig:ps}). This suggests that pruning tolerance is governed more by model architecture design than by parameter count alone. Hybrid linear attention MoE models are more robust under extreme compression than standard MoE (\autoref{tab:r}).


\textbf{Finding 3: Average accuracy masks workload and model-specific failures.}
Average quality alone is insufficient to characterize compressed MoEs: Prune+Quant retains high average performance ($\mathrm{Score}_{\mathrm{perf}}=96.35$) but drops to $\mathrm{Score}_{\mathrm{robust}}=88.91$, indicating uneven degradation across workloads  (\autoref{tab:ae}). The largest drop is seen on MMLU and HumanEval for Qwen3-30B-A3B, while GSM8K, MuSR, and RULER remain comparatively stable; similarly, coder-oriented variants show larger PPL inflation than instruction-tuned models under comparable pruning (\autoref{tab:r}). 
This motivates reporting workload robustness alongside average retention.

\textbf{Finding 4: KV-cache compression is robust at 8-bit but sensitive at lower precision.}
KV-q8 preserves short-context accuracy across baseline, quantized, pruned, and composable settings, while q4 and TurboQuant-style variants cause larger losses (\autoref{tab:kvcs}). Although smaller than expert-pruning loss, KV degradation grows at lower precision and is more pronounced in pruned models. 
Maintaining higher precision for key tensors preserves quality.

\noindent\textbf{Finding 5: At the same memory target, different compression combinations are not equivalent.}
\autoref{fig:ci} shows that composable compression losses are not additive: at $\sim$75--80\% compression, Q4-based prune+quant remains near the quantization-only PPL, while IQ1\_M sharply increases PPL despite similar or larger size reduction. Adding KV-q8 changes PPL only marginally, so expert pruning and low-bit weight precision dominate quality loss, while KV compression mainly adds memory savings.


\section{Efficiency Evaluation}
\begin{figure}
    \centering
    \includegraphics[width=\linewidth]{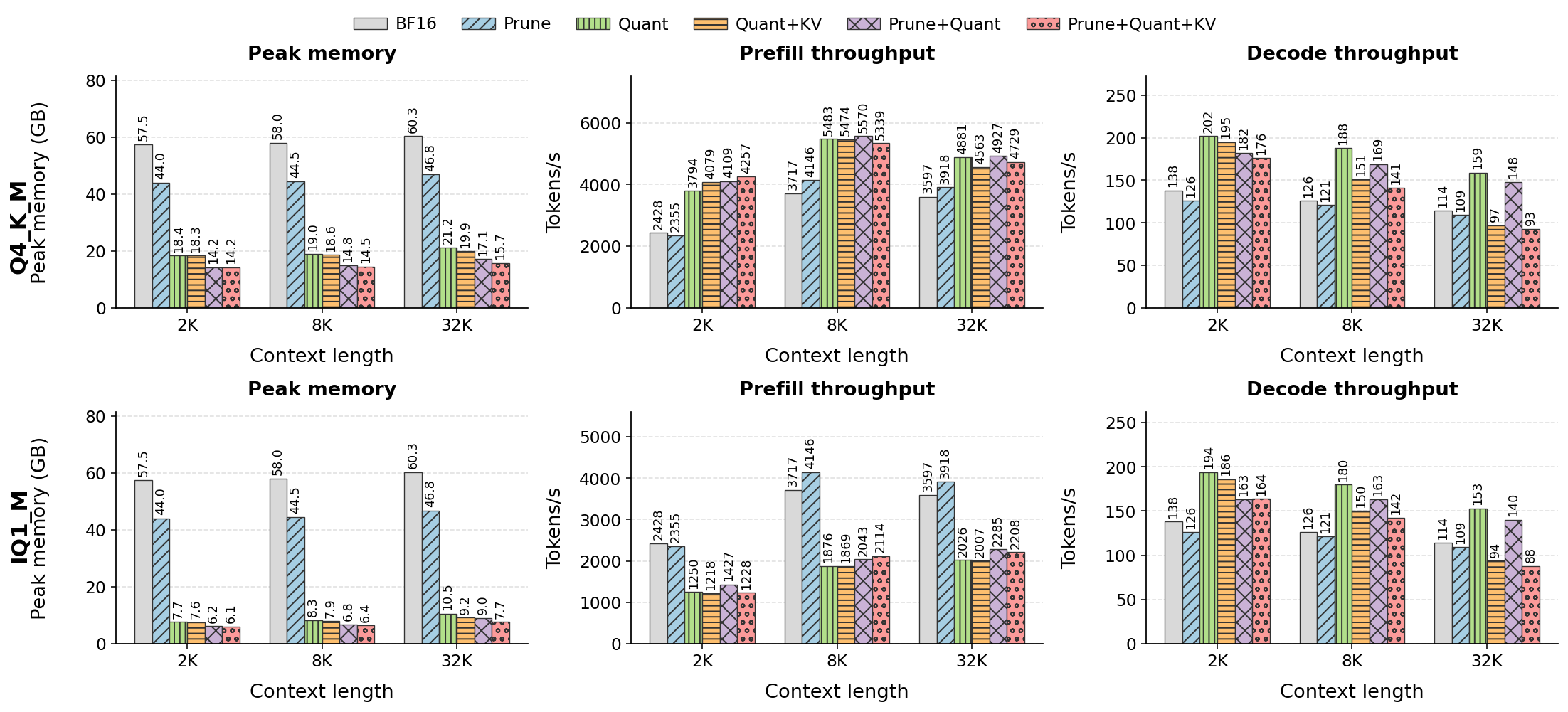}
    \caption{Inference efficiency.
Peak memory, prefill throughput, and decode throughput on NVIDIA~1xH100 for Qwen3-30B-A3B-Instruct at varying context lengths, across the six compression configurations for Q4\_K\_M (top) and IQ1\_M (bottom) with Q8 KV-cache under bs=1. Both higher compression rate and higher throughput are desirable although their relationship is non-monotonic.
}
    \label{fig:ie}
\end{figure}
\begin{figure}
    \centering
    \begin{minipage}[b]{0.48\linewidth}
        \centering
        \includegraphics[width=\linewidth,height=4.2cm,keepaspectratio]{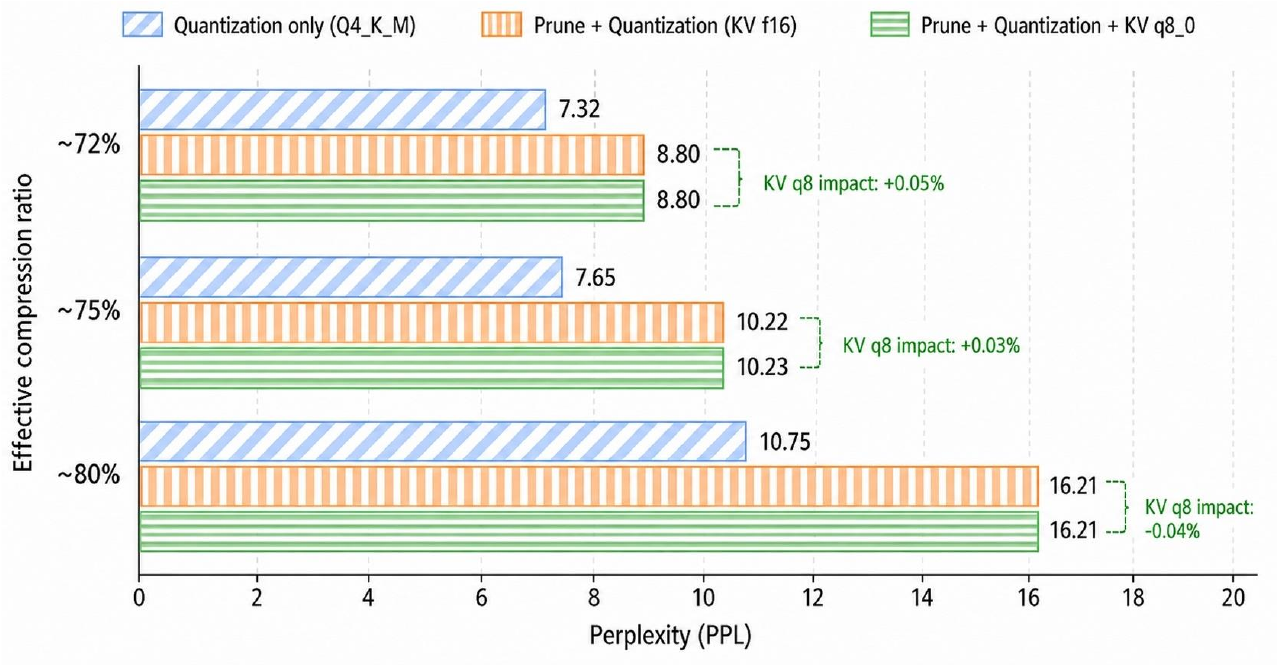}
        \caption{
        Performance under extreme compression scenario on Qwen3.6-35B-A3B. Similar compression rates yield different PPL.
        }
        \label{fig:ci}
    \end{minipage}
    \hfill
    \begin{minipage}[b]{0.48\linewidth}
        \centering
        \includegraphics[width=\linewidth,height=4.2cm,keepaspectratio]{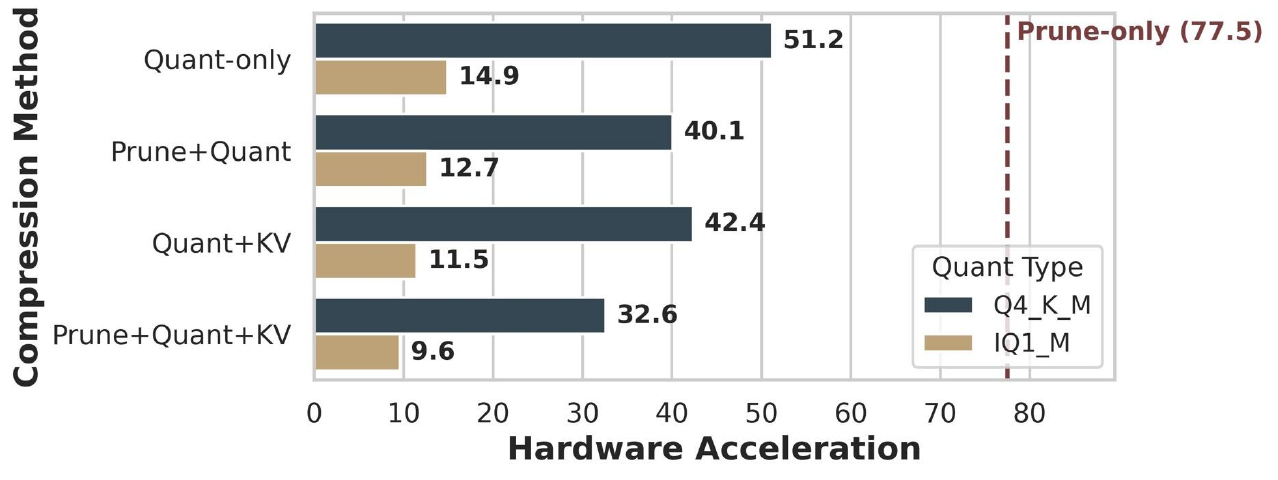}
        \caption{Hardware acceleration on H100 at 32K context length. Effective memory compression ratio vs.\ absolute prefill speedup; the gap shows how much theoretical compression fails to translate into runtime.
        }
        \label{fig:hw}
    \end{minipage}
\end{figure}

We answer RQ3 using Modules 7-8 to test whether compression reduces practical deployment cost.

\noindent\textbf{Finding 6: Smaller memory footprint does not guarantee higher throughput.}
We challenge the common assumption that memory reduction translates monotonically into faster inference. 
Q4\_K\_M Prune+Quant cuts peak memory from 60.3 GB to 17.1\,GB while improving prefill by $1.37\times$ and decode by $1.30\times$ over BF16 in \autoref{fig:ie}. Yet adding KV compression further reduces memory to 15.7\,GB but lowers prefill to $0.96\times$ and decode to $0.63\times$ of Prune+Quant. The effect is phase and context-dependent: Q4 Quant+KV nearly preserves 2K decode ($0.97\times$) but reduces 32K decode to $0.61\times$ of Quant-only. Quantization scheme also matters: IQ1\_M uses less memory than Q4\_K\_M at 32K Quant-only (10.5 GB vs. 21.2\,GB), but achieves only $0.41\times$ the Q4\_K\_M prefill throughput, showing that aggressive low-bit dequantization overhead can dominate memory savings.




\noindent\textbf{Finding 7: KV compression reduces peak memory, but its runtime benefit deteriorates at long context.} We challenge the hypothesis that KV compression should improve runtime by reducing memory traffic.
\autoref{fig:ie} demonstrates that at Q4\_K\_M (32K context), adding KV compression to Quant-only reduces peak memory from 21.2 to 19.9\,GB, but decode drops to $0.61\times$ while prefill stays at $0.93\times$. The same holds after pruning: Prune+Quant+KV lowers memory from 17.1 to 15.7\,GB, but decode falls to $0.63\times$ while prefill remains $0.96\times$. This penalty is context-sensitive: Q4 Quant+KV preserves 2K decode at $0.97\times$, but falls to $0.61\times$ at 32K, since long-context decode repeatedly reads and dequantizes the compressed KV cache.
This trend is also reflected on M1 Max, where KV-q8 at 32K reduces peak memory by 46\% but also lowers decode throughput by 46\%.

\noindent\textbf{Finding 8: Compression effects on throughput are hardware-dependent.}
Compression reduces memory on both H100 and M1 Max, but the throughput benefit is not proportional. At 32K, Q4\_K\_M on H100 cuts peak memory from 60.3 to 21.2\,GB and improves prefill by $1.36\times$ over BF16 (\autoref{fig:ie}). On M1 Max (\autoref{sec:appendix-m1max}), Q4\_K\_M also reduces estimated total memory, from roughly 33.3 to 20.4\,GiB, but prefill remains near parity with Q8\_0 ($0.97\times$). Thus, memory savings are portable, but throughput gains depend on whether the runtime can hide dequantization and cache-format overhead.



\noindent\textbf{Finding 9: Expert pruning provides more effective hardware acceleration than aggressive low-bit quantization.}
We challenge the assumption that larger memory compression implies better hardware acceleration. As shown in \autoref{fig:hw}, prune-only achieves smaller compression on H100 ($\sim$1.3x) but preserves the BF16 execution path and maintains near-baseline speedup (0.94-1.03), yielding Score$_\mathrm{hw}=72.0$-$79.4$. In contrast, Q4\_K\_M compresses more (2.8x-3.1x) but achieves only 1.38x-1.51x effective speedup, reducing Score$_\mathrm{hw}$ to $\sim$48; IQ1\_M falls further to 8.6x-12.8x as unpacking and dequantization overhead dominate.

\vspace{-2ex}
\section{Discussion}
\label{sec:discuss}
\vspace{-2ex}

\textbf{Compression Interaction.} Comparing the six radar plots in \autoref{fig:o} shows that composable compression mainly shifts sensitivity and hardware axes, not average accuracy. Quantization preserves $S_{\mathrm{perf}}$ and $S_{\mathrm{robust}}$ near baseline, while adding pruning drives $S_{\mathrm{arch}}$ toward the pruning-only regime, indicating pruning-dominated reliability loss. KV compression has little quality impact but further reduces $S_{\mathrm{hw}}$. Thus, the main harmful interaction is not additive accuracy degradation, but reduced cross-architecture reliability and hardware efficiency under composable compression.



\textbf{Practical Guidelines for MoE Compression.}
Our results suggest several deployment guidelines. 
First, practitioners should optimize for the target accuracy-memory-latency point rather than maximum size reduction, since smaller artifacts do not necessarily improve throughput. Second, moderate-bit weight quantization is often a low-risk first step, whereas expert pruning requires architecture-specific tuning. Third, KV-cache compression should be treated primarily as a long-context memory reduction technique, not as a guaranteed runtime optimization, since on-the-fly KV dequantization can slow decode at long context. Finally, composable compression should be validated as a complete deployment pipeline on the target backend: average accuracy, workload robustness, architecture reliability, peak memory, prefill throughput, and decode throughput should be reported together before selecting a configuration.


\textbf{Why a separate MoE specific benchmark?}
(1) \textit{MoE introduces a unique compression axis}: Expert pruning reduces memory but not active computation, so smaller MoE models may see little or no speedup.
(2) \textit{Expert compression dominates and interacts with other compression}: It causes far greater quality loss than quantization and amplifies the penalties of both weight and KV-cache compression, making results dependent on the pruning ratio.
(3) \textit{Results do not transfer across MoE models}: a single-model study cannot substitute for a benchmark.
(4) \textit{Existing benchmarks do not capture joint MoE compression}: A proper benchmark must evaluate pruning, quantization, and KV compression together across multiple MoE families with both quality and speed metrics.

\textbf{Does accuracy depend on the order of compression?}
It has limited order sensitivity (but is not order invariant) as the compression techniques are applied on largely independent axes (model weights, KV cache, sparsity). KV-cache compression is a runtime setting invariant to offline compression schemes. Ref. \autoref{sec:appendix-ablation}.

\textbf{Limitations.}
We use representative methods for each compression axis and leave alternatives such as expert parameter sharing and non-GGUF quantization to future work. Additionally, we agree serving-layer techniques e.g. speculative decoding, concurrent batching or optimized kernels would shift the latency picture; our efficiency claims are scoped to the evaluated execution path, and we report both CUDA and Metal to show how much findings vary across backends. Future work will also expand to more models and devices.

\section{Conclusion}
We present \sys{}, an end-to-end benchmark for MoE compression deployment. We envision our work as a milestone to guide the community in choosing and understanding the combination of compression strategies for deploying Mixture-of-Experts models in practice. We hope \sys{} can provide  insightful takeaways and findings for Mixture-of-Experts model compression design and serve as a solid foundation for future benchmarks. Our end-to-end model compression tool, including configuration and raw measurements, will be open-sourced upon acceptance.

\bibliographystyle{plainnat}
\bibliography{main}
\appendix
\section{Ablation Study}
\label{sec:appendix-ablation}

\subsection{Calibration Dataset}
\label{subsec:appendix-calibration}
\label{app:cal}

REAP estimates expert saliency from calibration routing statistics, so the calibration corpus can bias which capabilities are preserved after pruning. We isolate this effect on Qwen3.6-35B-A3B by fixing the pruning ratio (\(r=0.30\)), seed, router renormalization, and final GGUF Q4\_K\_M export, and varying only the calibration source.

\begin{table}[h]
\centering
\small
\caption{Calibration-dataset ablation for REAP on Qwen3.6-35B-A3B. All variants use the same pruning ratio \(r=0.30\), seed, router renormalization, and GGUF Q4\_K\_M deployment path. Higher is better for all metrics.}
\label{tab:calib_ablation}
\begin{tabular}{lccccccc}
\toprule
Calibration & GSM8K & MMLU-R & MuSR & HumanEval & MBPP & Math-Hard & IFEval \\
\midrule
Official mixture & 0.883 & 0.811 & 0.431 & 0.713 & 0.148 & 0.539 & 0.797 \\
C4 general       & \textbf{0.898} & \textbf{0.827} & 0.418 & 0.281 & \textbf{0.524} & 0.297 & 0.771 \\
Code-only        & 0.891 & 0.796 & 0.427 & 0.665 & 0.516 & 0.557 & 0.782 \\
Math-only        & 0.896 & 0.804 & \textbf{0.438} & \textbf{0.720} & 0.264 & \textbf{0.560} & \textbf{0.815} \\
\bottomrule
\end{tabular}
\end{table}

Table~\ref{tab:calib_ablation} shows that calibration materially changes the retained capability profile. C4 gives the best GSM8K and MMLU-R scores but is weakest on MuSR, HumanEval, Math-Hard, and IFEval; code-only calibration improves MBPP but not HumanEval; math-only calibration is strongest on MuSR, Math-Hard, HumanEval, and IFEval. Calibration can therefore change the interpretation of pruning results and should be treated as a first-class REAP hyperparameter. The official mixture is not uniformly optimal, but it remains the safest general-purpose default because it avoids the largest single-domain failures.

\subsection{Quantization-aware REAP}
\label{subsec:appendix-quant-aware-reap}

REAP collects saliency scores before pruning, while the deployed artifact is pruned, exported to GGUF, and quantized. This can create a score--deployment mismatch if low-bit quantization changes expert usefulness. We test a 4-bit packed-MoE-aware scoring surrogate used only for expert ranking; the selected pruning mask is still applied to the original full-precision checkpoint before GGUF Q4\_K\_M export.

\begin{table}[h]
\centering
\small
\caption{Quantization-aware REAP ablation on Qwen3.6-35B-A3B. All pruned variants use REAP with $r=0.30$ and are evaluated after GGUF Q4\_K\_M quantization through the same deployment path. ``Module'' quantizes only standard linear modules during scoring, while ``Packed MoE'' also covers packed routed experts and router weights.}
\label{tab:quant_aware_reap}
\begin{tabular}{l c c c}
\toprule
Scoring setup & GSM8K & MMLU-R & MuSR \\
\midrule
Unpruned Q4\_K\_M
& \textbf{0.8916}
& \textbf{0.8625}
& \textbf{0.4365} \\

BF16 REAP
& \textbf{0.8825}
& 0.8111
& \textbf{0.4312} \\

Module quant.-aware REAP
& 0.8741
& 0.8107
& 0.4272 \\

Packed-MoE quant.-aware REAP
& 0.8779
& \textbf{0.8131}
& 0.4299 \\
\bottomrule
\end{tabular}
\end{table}

Table~\ref{tab:quant_aware_reap} shows that packed-MoE-aware scoring improves consistently over module-level quantization-aware scoring, but not over standard BF16 REAP. For Qwen3.6-35B-A3B at $r=0.30$, the remaining degradation is better explained by structural expert removal than by BF16-to-GGUF mismatch, so BF16 REAP remains the default pruning path.

\subsection{Alternate expert compression methods}
\label{subsec:appendix-ream}
Although our original scope selected one representative method per compression axis, we acknowledge that a single method can be deemed as insufficient to establish method-independent conclusions. We add REAM \cite{jha2026ream}, a representative expert-merging method, and compare it with REAP at a matched 25\% expert-count reduction on the seven Table 3 workloads, using the same backend and evaluation protocol.

First, the expert-compression axis remains the dominant source of aggregate quality loss (finding 1). Starting from score\_perf=100, the expert-only stage causes loss of 5.31 points under REAP and 6.05 under REAM. Adding both Q4\_K\_M and Q8 KV contributes only 1.23 and 0.31 additional points, respectively. The fully composable configurations also retain nearly identical average performance (93.46 vs 93.64). This trend is expected because both pruning and merging directly alter expert capacity and specialization, whereas weight and KV cache quantization preserve the model structure and introduce only lower-precision representations. Thus, expert-axis dominance is not specific to REAP.

Second, the qualitative trend across compression axes is consistent across both expert-compression methods tested, while the exact magnitudes can slightly differ. Adding Q4\_K\_M changes score\_perf by -1.18 under REAP but -0.06 under REAM, while adding Q8 KV changes it by -0.05 and -0.25, respectively. This also reconfirms finding 5 (isolated results cannot fully predict composable configurations.)

We include the complete per-workload results rather than claim identical/exact behaviour across methods in \autoref{tab:ream}.

\begin{table*}[t]
\centering
\caption{Performance comparison of expert compression methods across configurations.}
\label{tab:compression_results}
\resizebox{\textwidth}{!}{%
\begin{tabular}{llcccccccccc}
\toprule
\textbf{Expert comp. method}
& \textbf{Config}
& \textbf{GSM8K}
& \textbf{MMLU}
& \textbf{MuSR}
& \textbf{IFEval}
& \textbf{HumanEval}
& \textbf{BFCLv3}
& \textbf{RULER}
& \textbf{Avg.}
& \textbf{Score\_perf}
& \textbf{Score\_robust} \\
\midrule

--- & Baseline BF16
& 0.8360 & 0.8470 & 0.4290 & 0.8170
& 0.9390 & 0.8880 & 0.9441 & 0.8143
& 100.00 & 100.00 \\

REAP & Expert only
& 0.8400 & 0.7420 & 0.4380 & 0.7650
& 0.8110 & 0.8560 & 0.9452 & 0.7710
& 94.69 & 80.41 \\

--- & Q4\_K\_M only
& 0.8330 & 0.8450 & 0.4250 & 0.8260
& 0.9270 & 0.8910 & 0.9422 & 0.8127
& 99.81 & 97.43 \\

--- & Q4\_K\_M + Q8 KV
& 0.8378 & 0.8422 & 0.4246 & 0.8336
& 0.9268 & 0.8694 & 0.9435 & 0.8111
& 99.61 & 95.56 \\

REAP & + Q4\_K\_M
& 0.8230 & 0.7330 & 0.4310 & 0.7430
& 0.8170 & 0.8410 & 0.9424 & 0.7615
& 93.51 & 80.82 \\

REAP & + Q4\_K\_M + Q8 KV
& 0.8180 & 0.7320 & 0.4270 & 0.7600
& 0.8050 & 0.8420 & 0.9435 & 0.7611
& 93.46 & 80.41 \\

\midrule

\textbf{REAM} & \textbf{Expert only}
& \textbf{0.8658} & \textbf{0.7377} & \textbf{0.4246} & \textbf{0.7634}
& \textbf{0.7317} & \textbf{0.8810} & \textbf{0.9509} & \textbf{0.7650}
& \textbf{93.95} & \textbf{70.74} \\

\textbf{REAM} & \textbf{+ Q4\_K\_M}
& \textbf{0.8673} & \textbf{0.7321} & \textbf{0.4233} & \textbf{0.7745}
& \textbf{0.7134} & \textbf{0.8915} & \textbf{0.9497} & \textbf{0.7645}
& \textbf{93.89} & \textbf{68.76} \\

\textbf{REAM} & \textbf{+ Q4\_K\_M + Q8 KV}
& \textbf{0.8628} & \textbf{0.7351} & \textbf{0.4193} & \textbf{0.7579}
& \textbf{0.7256} & \textbf{0.8875} & \textbf{0.9497} & \textbf{0.7625}
& \textbf{93.64} & \textbf{70.11} \\

\bottomrule
\label{tab:ream}
\end{tabular}%
}
\end{table*}
These experiments extend the evidence from pruning to merging and show that the central claims: expert-axis dominance and non-separable composable effects are not artifacts of REAP. It is to be noted that we do not claim method independence over all possible algorithms.

\section{Sensitivity Studies}

\subsection{Pruning Sensitivity}
\label{sec:appendix-prune}

\begin{table}[t]
\centering
\caption{Pruning sensitivity under fixed quantization and KV-cache settings. For each pruning ratio, the bf16/f16 row serves as the same-prune baseline.}
\label{tab:pruning_sensitivity}
\resizebox{\linewidth}{!}{
\begin{tabular}{c c c c c c}
\toprule
\textbf{Pruning} &
\textbf{Weights} &
\textbf{KV cache} &
\textbf{PPL $\pm$ stderr} &
\textbf{$\Delta$ vs 0\% bf16} &
\textbf{$\Delta$ vs same-prune bf16} \\
\midrule

0\%  & bf16    & f16   & $6.8200$              & --                    & -- \\
0\%  & Q4\_K\_M & f16   & $6.8198 \pm 0.0444$   & $-0.0002$             & $-0.0002$ \\
0\%  & Q4\_K\_M & q8\_0 & $6.8180 \pm 0.0444$   & $-0.0020$             & $-0.0020$ \\

\midrule

20\% & bf16    & f16   & $8.6695 \pm 0.0592$   & $+1.8495$             & -- \\
20\% & Q4\_K\_M & f16   & $8.7952 \pm 0.0602$   & $+1.9752$             & $+0.1257$ \\
20\% & Q4\_K\_M & q8\_0 & $8.7998 \pm 0.0602$   & $+1.9798$             & $+0.1303$ \\

\midrule

30\% & bf16    & f16   & $10.0449 \pm 0.0706$  & $+3.2249$             & -- \\
30\% & Q4\_K\_M & f16   & $10.2242 \pm 0.0721$  & $+3.4042$             & $+0.1793$ \\
30\% & Q4\_K\_M & q8\_0 & $10.2272 \pm 0.0721$  & $+3.4072$             & $+0.1823$ \\

\midrule

50\% & bf16    & f16   & $15.8121 \pm 0.1206$  & $+8.9921$             & -- \\
50\% & Q4\_K\_M & f16   & $16.2116 \pm 0.1246$  & $+9.3916$             & $+0.3995$ \\
50\% & Q4\_K\_M & q8\_0 & $16.2051 \pm 0.1245$  & $+9.3851$             & $+0.3930$ \\

\bottomrule
\end{tabular}
}
\end{table}

As shown in \autoref{tab:pruning_sensitivity}, across 0\%--50\% pruning, the same-prune gaps from Q4\_K\_M weights and q8\_0 KV cache remain small relative to the pruning-induced PPL increase. Pruning is therefore the dominant source of degradation in this sweep.

\subsection{Quantization Sensitivity}
\label{sec:appendix-quant}

We report the without-IQ1\_M version of $\mathrm{Score}_{\mathrm{quant}}$ as the primary radar score because IQ1\_M is a deployment-uncommon outlier. As shown in \autoref{tab:quant_score_iq1}, including IQ1\_M lowers the stability term to roughly 0.60 and compresses all scores into the 53--56 range; excluding it raises the scores by about 30 points and better reflects the practical quantization regime.

\begin{table}[h]
\centering
\small
\caption{Quantization sensitivity score with and without IQ1\_M.}
\label{tab:quant_sensitivity_iq1m}
\begin{tabular}{lcc}
\hline
Method & Score (with IQ1\_M) & Score (without IQ1\_M) \\
\hline
Quant-only            & 56.12 & 87.45 \\
Quant+KV (q8)         & 56.10 & 87.42 \\
Prune+Quant           & 55.67 & 84.35 \\
Prune+Quant+KV (q8)   & 55.62 & 84.32 \\
Prune+Quant+KV (q4)   & 53.21 & 83.14 \\
\hline
\end{tabular}
\end{table}

\section{Inference Efficiency on Apple M1 Max}
\label{sec:appendix-m1max}

\begin{figure}[h]
    \centering
    \includegraphics[width=\linewidth]{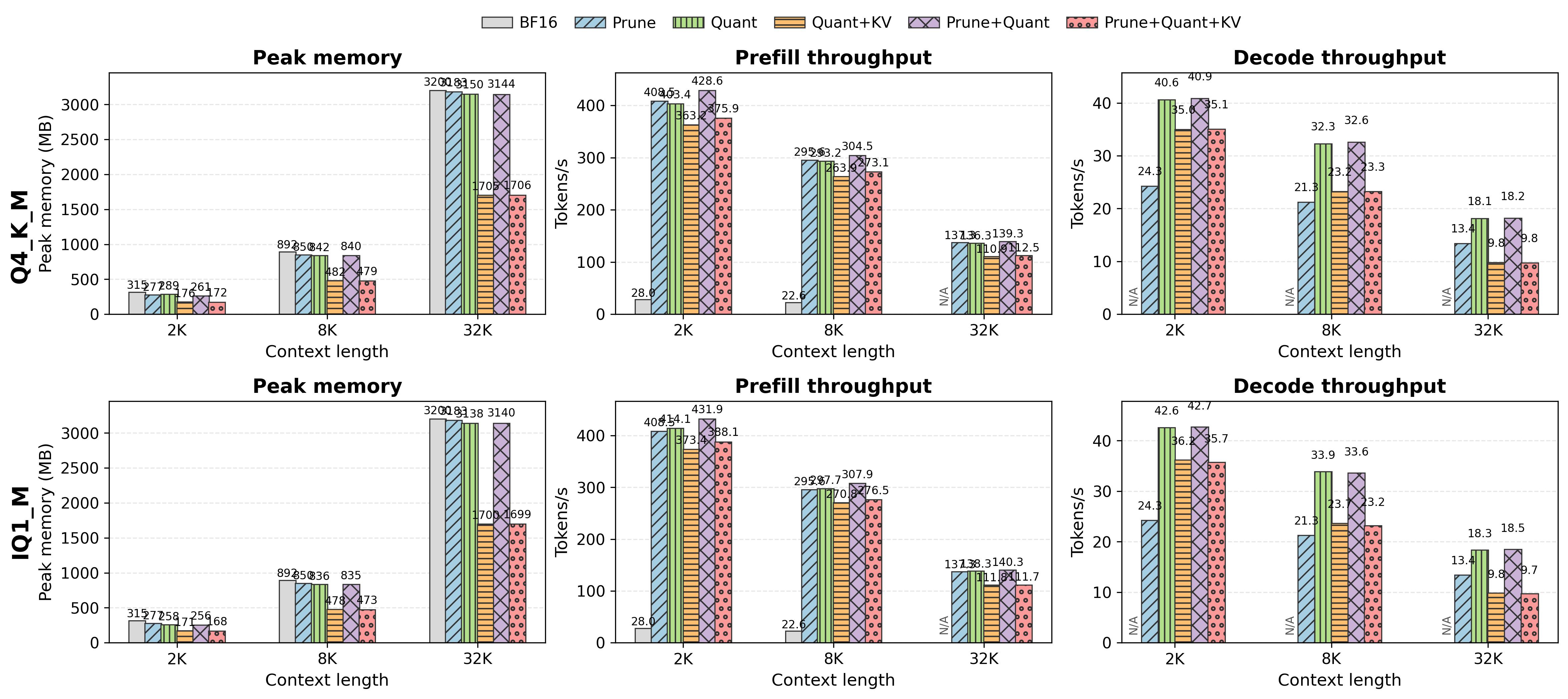}
    \caption{
    Inference efficiency on Apple-M1-Max. Peak memory, prefill throughput, and decode throughput for Qwen3-30B-A3B-Instruct at context lengths $\{2\mathrm{K}, 8\mathrm{K}, 32\mathrm{K}\}$, across the six compression configurations of \autoref{tab:compression_configs} at two weight-quantization formats (Q4\_K\_M, IQ1\_M) with Q8 KV-cache. Q8\_0 is used as the baseline because BF16 caused memory thrashing for this model.
    }
    \label{fig:m1max}
\end{figure}

Figure~\ref{fig:m1max} reports peak memory, prefill throughput, and decode throughput on Apple~M1~Max for Qwen3-30B-A3B-Instruct at context lengths $\{2\mathrm{K}, 8\mathrm{K}, 32\mathrm{K}\}$, across the same six compression configurations of \autoref{tab:compression_configs} at Q4\_K\_M and IQ1\_M weight quantization with Q8 KV-cache. In this unified-memory regime, long-context footprint is dominated by KV cache and runtime buffers rather than weights. As a result, weight-only compression yields limited memory benefit, whereas KV compression lowers peak memory substantially but also hurts decode throughput because on-the-fly dequantization stays on the critical path.

\section{Reproducibility}
All model checkpoints used in our experiments are listed in \autoref{tab:checkpoints}.

\begin{table*}[t]
\centering
\scriptsize
\setlength{\tabcolsep}{2.5pt}
\renewcommand{\arraystretch}{0.88}
\caption{Original and REAP-pruned Hugging Face checkpoints.}
\label{tab:hf_reap_ids}
\resizebox{\textwidth}{!}{
\begin{tabular}{@{}lll@{}}
\toprule
\textbf{Model} & \textbf{Original HF ID} & \textbf{REAP-pruned HF ID} \\
\midrule
Qwen3-30B-A3B &
\path{Qwen/Qwen3-30B-A3B-Instruct-2507} &
\path{SamsungSAILMontreal/Qwen3-30B-A3B-Instruct-2507-REAP} \\

Qwen3-Coder-30B-A3B &
\path{Qwen/Qwen3-Coder-30B-A3B-Instruct} &
\path{cerebras/Qwen3-Coder-REAP-25B-A3B} \\

Qwen3-235B-A22B &
\path{Qwen/Qwen3-235B-A22B-Instruct-2507} &
\path{SamsungSAILMontreal/Qwen3-235B-A22B-Instruct-2507-REAP} \\

GLM-4.7-Flash &
\path{zai-org/GLM-4.7-Flash} &
\path{cerebras/GLM-4.7-Flash-REAP-23B-A3B} \\

MiniMax-M2 &
\path{MiniMaxAI/MiniMax-M2} &
\path{cerebras/MiniMax-M2-REAP-162B-A10B} \\

Kimi-Linear-48B-A3B &
\path{moonshotai/Kimi-Linear-48B-A3B-Instruct} &
\path{cerebras/Kimi-Linear-REAP-35B-A3B-Instruct} \\

Qwen3-Next-80B-A3B &
\path{Qwen/Qwen3-Next-80B-A3B-Instruct} &
\path{SamsungSAILMontreal/Qwen3-Next-80B-A3B-Instruct-REAP} \\

Qwen3-Coder-Next &
\path{Qwen/Qwen3-Coder-Next} &
\path{SamsungSAILMontreal/Qwen3-Coder-Next-REAP} \\

Qwen3.6-35B-A3B &
\path{Qwen/Qwen3.6-35B-A3B} &
\path{RangerX/Qwen3.6-35B-REAP-Pruned-ratio-0.2} \\

Step-3.5-Flash &
\path{stepfun-ai/Step-3.5-Flash} &
\path{cerebras/Step-3.5-Flash-REAP-149B-A11B} \\
\bottomrule
\label{tab:checkpoints}
\end{tabular}
}
\end{table*}
\end{document}